\documentclass[9pt]{IEEEtran}

\newcommand{\subparagraph}{}
\usepackage[compact]{titlesec}
\usepackage{graphicx}
\usepackage{amsmath}
\usepackage{amssymb}
\usepackage{algorithmic}
\usepackage{subcaption}
\usepackage{stfloats}
\usepackage{float}
\usepackage{url}
\usepackage[font = footnotesize]{caption}
\usepackage{xcolor}
\usepackage[linesnumbered,ruled,vlined]{algorithm2e}
\usepackage{fancyhdr}

\begin{document}

\bstctlcite{IEEEexample:BSTcontrol}

\title{PT-Spike: A Precise-Time-Dependent Single Spike Neuromorphic Architecture with Efficient Supervised Learning}

\author{
\IEEEauthorblockN{Tao Liu\IEEEauthorrefmark{1}, Lei Jiang\IEEEauthorrefmark{2}, Yier Jin\IEEEauthorrefmark{3}, Gang Quan\IEEEauthorrefmark{1} and Wujie Wen\IEEEauthorrefmark{1}}


\IEEEauthorblockA{\IEEEauthorrefmark{1}\textit{Florida International University}, \IEEEauthorrefmark{2}\textit{Indiana University}, 
\IEEEauthorrefmark{3}\textit{University of Florida}\\
\IEEEauthorrefmark{1}\{tliu023, gang.quan, wwen\}@fiu.edu, \IEEEauthorrefmark{2}jiang60@iu.edu, \IEEEauthorrefmark{3}yier.jin@ece.ufl.edu}
\vspace{-25pt}

\thanks{This work is supported in part by NSF under project CNS-1423137 and 2016-2017 Collaborative Seed Award Program of Florida Center for Cybersecurity (FC$^2$).}


}

\maketitle
\thispagestyle{fancy}
\lhead{}\chead{}\rhead{}
\lfoot{}\cfoot{}\rfoot{}
\renewcommand{\headrulewidth}{0pt}
\renewcommand{\footrulewidth}{0pt}
\pagestyle{fancy}
\begin{abstract}
One of the most exciting advancements in Artificial Intelligence (AI) over the last decade is the wide adoption of Artificial Neural Networks (ANNs), such as Deep Neural Network (DNN) and Convolutional Neural Network (CNN), in real world applications. However, the underlying massive amounts of computation and storage requirement greatly challenge their applicability in resource-limited platforms like drone, mobile phone and IoT devices etc. The third generation of neural network model--Spiking Neural Network (SNN), inspired by the working mechanism and efficiency of human brain, has emerged as a promising solution for achieving more impressive computing and power efficiency within light-weighted devices (e.g. single chip). However, the relevant research activities have been narrowly carried out on conventional rate-based spiking system designs for fulfilling the practical cognitive tasks, underestimating SNN's energy efficiency, throughput and system flexibility. Although the time-based SNN can be more attractive conceptually, its potentials are not unleashed in realistic applications due to lack of efficient coding and practical learning schemes. In this work, a \textit{P}recise-\textit{T}ime-Dependent Single \textit{Spike} Neuromorphic Architecture, namely \textit{``PT-Spike"}, is developed to bridge this gap. Three constituent hardware-favorable techniques: 
precise single-spike temporal encoding,
efficient 
supervised temporal learning and 
fast 
asymmetric decoding are proposed accordingly to boost the energy efficiency and data processing capability of the time-based SNN at a more compact neural network model size when executing real cognitive tasks.
Simulation results show that \textit{``PT-Spike"} demonstrates significant improvements in network size, processing efficiency and power consumption with marginal classification accuracy degradation, when compared with the rate-based SNN and ANN under the similar network configuration.
\end{abstract}

\section{Introduction}
\label{sec:intro}

Deep learning enabled neural network system, i.e. deep neural network (DNN) or convolutional neural network (CNN), has found broad applications in realistic cognitive tasks such as speech recognition, image processing, machine translation and object detection ~\cite{lecun2015deep,szegedy2016overview}. 
However, performing high-accurate testings for complex DNNs or CNNs requires massive amounts of computation and memory resources, leading to limited energy efficiency. For instance, the recognition implementation of 
CNN--AlextNet~\cite{krizhevsky2012imagenet} involves not only huge volumes of parameters (61 million) generating intensive off-chip memory accesses but also a large number of computing-intensive high precision floating-point operations (1.5 billion)~\cite{farmahini2015nda}. 
Such a weakness makes these solutions less attractive for many emerging applications of mobile autonomous systems like smart device, Internet-of-Things (IoT), wearable device, robotics etc., where very tighten power budget, hardware resource and footprint are enforced~\cite{andri2016yodann,han2016mcdnn}. 

Different from the CNN and DNN designs, spiking-based neuromorphic computing, which is inspired 
from the biological spiking neural network (SNN), has featured as achieving tremendous computing efficiency at much lower power of small footprint platforms, e.g. the famous IBM TrueNorth chip that has total 1 million synapses and an operating power of $\sim$70mW~\cite{akopyan2015truenorth}. These low-power, light, and small single-chip solutions leverage the efficient event-driven concept to ease the computational load and enable possible cognitive applications in resource limited platforms, creating a very unique but promising branch of neuromorphic computing research~\cite{neil2014minitaur,corradi2015neuromorphic}.




In spiking neuromorphic systems, the information is usually conveyed by the occurrence frequency of spikes (rate coding) or their firing time (time coding). Compared to the rate-based SNN, the more biological plausible time-based SNN may offer better energy efficiency and system throughput~\cite{thorpe2001spike}, since theoretically the information can be flexibly embedded in the time (temporal) domain of short and sparse spikes instead of the spiking count represented by a group of dense spikes in rate coding, e.g. the spike occurrence frequency is proportional to the intensity of the input like each pixel density of the image~\cite{akopyan2015truenorth,liu2016memristor}.
As a result, the rate-based SNN is naturally more power-hungry than that of time-based SNN due to the increased number of spikes and relevant spike operations, such as synaptic weighting and Integrate-and-Fire (IFC) etc.
Meanwhile, the processing efficiency of time-based SNN can be further enhanced by performing an early decision making based on the temporal information extracted from early fired spikes, while in rate coding, the classification cannot be initiated until the last moment, e.g. winner-takes-all rule by sorting the number of spikes fired during the entire period of decoding time for each output neuron~\cite{maass2000computational}.

However, the potentials of such an emerging architecture are significantly underestimated due to  lack of efficient hardware-favorable solutions for time-based information representation and complex spike-timing-dependent (temporal) training of biological synapses towards practical cognitive applications~\cite{wang2015energy}.
On one hand, translating the input stimulus (i.e. image pixels) to the delay of the spikes, namely time-based encoding, is non-trivial because the coding efficiency can be easily degraded by the biased spike delays distributed in the limited coding intervals. 
Also, the hardware realization of time coding is usually expensive, as the time-based spike kernel needs to be carefully designed to provide accurate time information (e.g. pre-synaptic/post-synaptic time~\cite{thorpe2001spike}) for time-based training.
On the other hand, realizing more biological plausible spiking-time based training, i.e. unsupervised spiking-time-dependent plasticity (STDP), is very complex and costly due to the exponential time dependence of weight change and difficult convergence of learning~\cite{sjostrom2010spike}. In real-world applications, training of the rate-based SNN can be usually performed off-line by directly borrowing the standard back-propagation algorithm from artificial neural network (ANN)~\cite{liu2016memristor}. However, this time-independent learning rule does not fit the time-dependent SNN because of a fundamentally different learning mechanism.

In this work, we investigate the possibility of unleashing the potentials of time-based \textbf{single-spike} SNN architecture in realistic applications by orchestrating the efficient time-based coding/decoding and learning algorithm. A \textit{\underline{P}recise-\underline{T}ime-Dependent Single \underline{Spike} Neuromorphic Architecture}, namely ``\textit{PT-Spike}", is proposed to facilitate the cognitive tasks like the MNIST digit recognition. Our ``\textit{PT-Spike}" incorporates three integrated techniques: precise single-spike temporal encoding, efficient supervised temporal learning, and fast asymmetric decoding.
Our major contributions are:
\begin{enumerate}
\item We develop a precise-temporal encoding approach to efficiently translate the information into the temporal domain of a single spike. The single spike solution dramatically reduces the energy, while offering efficient model size reduction;
\item We propose a supervised temporal learning algorithm to facilitate synaptic plasticity on this single-spike system. The proposed algorithm significantly improves the learning capability and achieves comparable accuracy when compared to the ANN and rate-based SNN under the similar configuration;

\item We design a novel asymmetric decoding to relieve the
unique and serious weight competition issue existing in this single-spike system, and significantly improve the efficacy and efficiency of synaptic weight updating.
\end{enumerate}


\section{Backgrounds and Motivations}
\label{sec:prelim}

\begin{figure}[t]
\begin{centering}
\includegraphics[width=1\columnwidth]{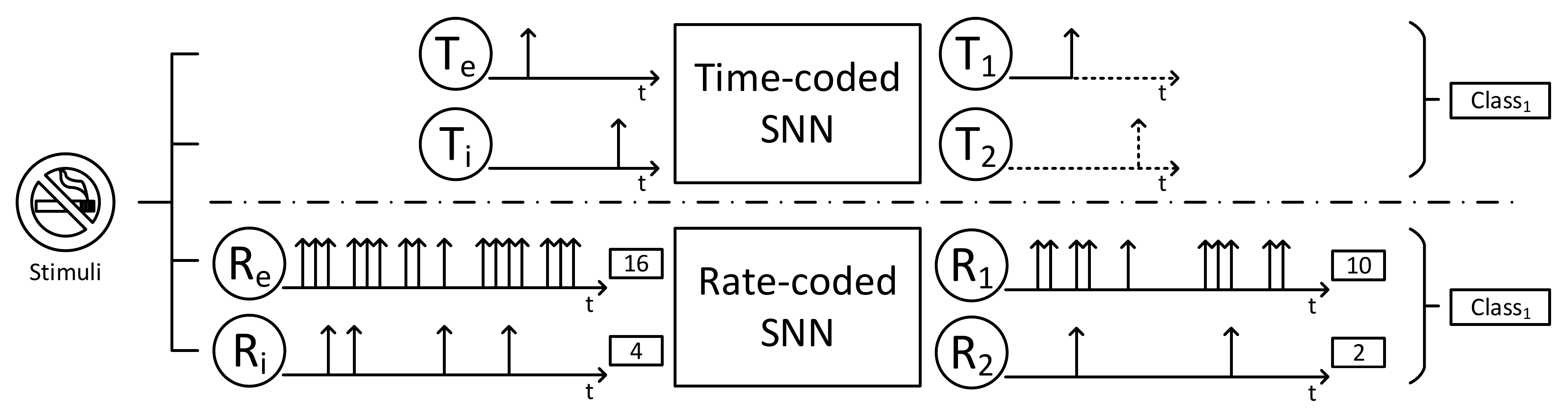}
\end{centering}
\vspace{-12pt}
\caption{%
\label{rdcode}
The Conceptual View of Rate-coding and Time-coding in SNNs. 
}
\vspace{-12pt}
\end{figure}

\subsection{Neural Coding in SNNs}
The neural coding in SNNs can be generally categorized as rate coding, time coding, rank coding and population coding etc.~\cite{borst1999information}. In particularly, the first two codings are the most attractive, since each piece of coded information is only associated with the spikes generated by a single input neuron, offering simplified encoding/decoding procedures and design complexity. 

Fig.~\ref{rdcode} demonstrates an example of conceptual comparison between rate coding and time coding in SNNs. $T{_e}$ and $T{_i}$ ($R{_e}$ and $R{_i}$) denote two types of input neurons: the time-coded (rate-coded) excitatory and inhibitory neurons, respectively. The excitatory neuron can exhibit an active response to the stimulus while the inhibitory neuron intends to keep silent. $T{_1}$ and $T{_2}$ ($R{_1}$ and $R{_2}$) denote two time-coded (rate-coded) output neurons for the classification. 
The rate-based SNN generates far more number of spikes than that of time-based SNN in both types of input neurons. 
After the input spikes are processed by the two different SNNs, a single spike firing at a specific time interval can perform an inference task in the output layer of the time-based SNN. However, a considerable number of spikes are needed for fulfilling a rate-based classification in the rate-based SNN, indicating a much higher power consumption. Moreover, the rate-based SNN may exhibit a slower processing speed than that of time-based SNN, since the output neuron of the former SNN needs to count the spiking numbers (i.e. through Integrate-and-Fire~\cite{burkitt2006review}) in the whole predefined time window, while that of the latter one may quickly suspend its computations once a spike is detected.

\subsection{Limitation of Existing Spiking Neuromorphic Computing Research}

\textbf{Neuromorphic Designs:} Many studies have been conducted to facilitate the spiking based Neuromorphic Computing System (NCS) designs in real hardware implementations, including CMOS VLSI circuit~\cite{akopyan2015truenorth,seo201145nm,cao2015spiking,esser2016convolutional}, reconfigurable FPGA~\cite{neil2014minitaur}, and emerging memristor crossbar~\cite{chu2015neuromorphic,liu2016memristor}. 
However, these works mainly focus on the rate- or time-based SNN model mapping and hardware implementations, rather than the SNN architecture optimization, i.e. coding, decoding and learning approaches etc. 

\textbf{Temporal Coding:} The concept of temporal coding, which relies on the arrival time or delay of a spike train for information representation, has been widely explored and proved in the development of time-based SNN~\cite{kempter1996temporal,butts2007temporal}. These theoretical studies, however, mainly emphasize on the biological explanations of time-based SNN models based on simple cognitive benchmarks (i.e. two inputs XOR gate), which are far from the complicated real-world problems such as image recognition. Recently, Zhao et al.~\cite{zhao2016energy} proposed an encoding circuit to handle the temporal coding, however, this type of work still concentrates on component-level hardware implementations with simple case studies, and hence is lack of a holistic architecture-level solution set capable of handling realistic tasks. In~\cite{yu2013precise}, a complete time-based SNN design is proposed. However, their solution suffers from limited accuracy fundamentally constrained by existing coding and temporal learning rule, and is not optimized towards hardware-based neuromorphic system designs.




\textbf{Temporal Learning:} Since the popular learning approaches such as back-propagation~\cite{rumelhart1988learning} widely used in ANN or rate-based SNN are unable to handle precise-time-dependent information due to a fundamentally different neural processing, many proposals dedicated to the time-based learning have been developed~\cite{sjostrom2010spike,gutig2006tempotron,ponulak2005resume}. However, these learning algorithms are neither hardware-favorable nor applicable for realistic tasks due to the expensive convergence and theoretical limitation. For example, in the unsupervised Spike-timing dependent plasticity (STDP) learning rule, the neural network structure and synaptic computation will be exponentially increased due to the expensive convergence and clustering. The proposed ``Tempotron" and ``Remote Supervised Method (ReSuMe)" can use the teaching spike to adjust desired spiking time for temporal learning, however, are not applicable to handle complicated patterns.

Our proposed \textit{``PT-Spike"} is substantially different from previous studies:
we explore how the time-based \textbf{single-spike} SNN architecture can be designed to perform the realistic tasks through a holistic efficient techniques spanning time-based coding, learning to decoding.
A low cost and efficient temporal learning named ``PT-Learning'' is augmented from the ``Tempotron'' learning by considering a synthesized contribution of the cost function and the hardware-favorable time-dependent kernel for weight updating. By integrating with proposed ``Precise Temporal Encoding'' and ``Asymmetric Decoding'', ``\textit{PT-Spike}'' can improve the accuracy, power, learning efficiency, and the model size reduction through the spatial-temporal information conversion significantly.

\section{Design Details}
\label{sec:design}

\begin{figure*}[t]
\begin{centering}
\includegraphics[width=1\textwidth]{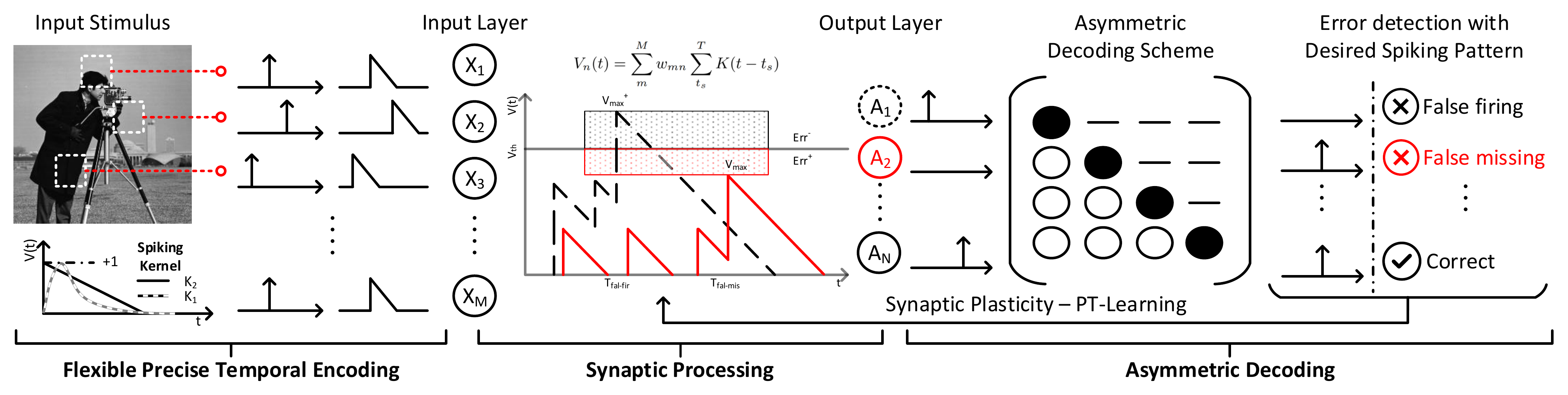}
\end{centering}
\vspace{-12pt}
\caption{%
\label{arch}
The overview of ``\textit{PT-Spike}" system architecture.
}
\vspace{-12pt}
\end{figure*}

\subsection{System Architecture}
Fig.~\ref{arch} shows a comprehensive data processing flow of proposed ``\textit{PT-Spike}". First, the stimulus will be captured by the temporal perceptors to generate a sparse spike train (i.e. \textbf{single spike}) through ``Precise Temporal Encoding". Each spike train will be further modulated in temporal domain by a linear-decayed spiking kernel to form time-dependent voltage pulse. 
Second, those voltage pulses will be sent to the synaptic network for a weighting process, i.e. the memristor crossbar with IFC design can be employed for parallel processing. The output neurons will exhibit time-varying weighting responses due to the time-dependent input information. After that, the output neuron will fire a spike if the weighted post-synaptic voltage crosses a threshold voltage. Then spike trains from the output layer will be transmitted to the ``Asymmetric Decoding". Finally, the target pattern will be classified by analyzing the synchronized output spikes with a predefined asymmetric rule. During the learning procedure, desired spike patterns are coded by following the similar asymmetric rule during decoding. The detected errors will be sent-back for synaptic plasticity through ``PT-Learning"--a supervised temporal learning algorithm.

\subsection{Precise Temporal Encoding}
As discussed in Section.~\ref{sec:prelim}, in traditional rate coding, a large number of spikes within a proper time window will be needed to precisely indicate the amplitude of an input signal, i.e. the pixel density of visual stimulus. To maximize the power efficiency with minimized number of spikes, the input information will be represented as an extreme sparse train--\textbf{single spike} and its occurring delay in aforementioned coding approach. However, such a ``one-to-one" mapping between each stimulus and spike train of each input neuron can lead to a significant energy overhead. Meanwhile, the time or temporal information of those spike trains are not fully leveraged by each neuron, resulting in limited coding efficiency thus a dramatical accuracy reduction. As we shall present later, our results on ``MNIST" benchmark show that the ``one-to-one'' mapping achieves very unacceptable training accuracy (($\sim20\%$) even under a large model size, that is, 784 input neurons for a $28\times28$ image. 

In ``\textit{PT-Spike}", we further propose the 
``Precise Temporal Encoding''.
As shown in Fig.~\ref{arch}, the ``Precise Temporal Encoding" is inspired from human visual cortex and Convolutional Neural Network (CNN), where a Temporal Kernel (i.e. a unit square matrix) will be applied on the full image to capture the spatial information and then translated into a single spike delay in temporal domain as a neuron input by perceiving the localized information from multiple interested pixels, i.e. spiking delay is equal to the average density among several selected pixels. In practice, by selecting a proper stride with which we slide the Temporal Kernel, e.g. smaller than the dimensionality of Temporal Kernel, a portion of localized spatial information will be shared by adjacent kernel sliding. Consequently, the spatial localities can be further transformed into temporal localities, thus to uniformly allocate the spiking delay assigned to each input neuron in time domain, translating into improved coding efficiency and classification accuracy.

\begin{figure}[b]
\begin{centering}
\includegraphics[width=1\columnwidth]{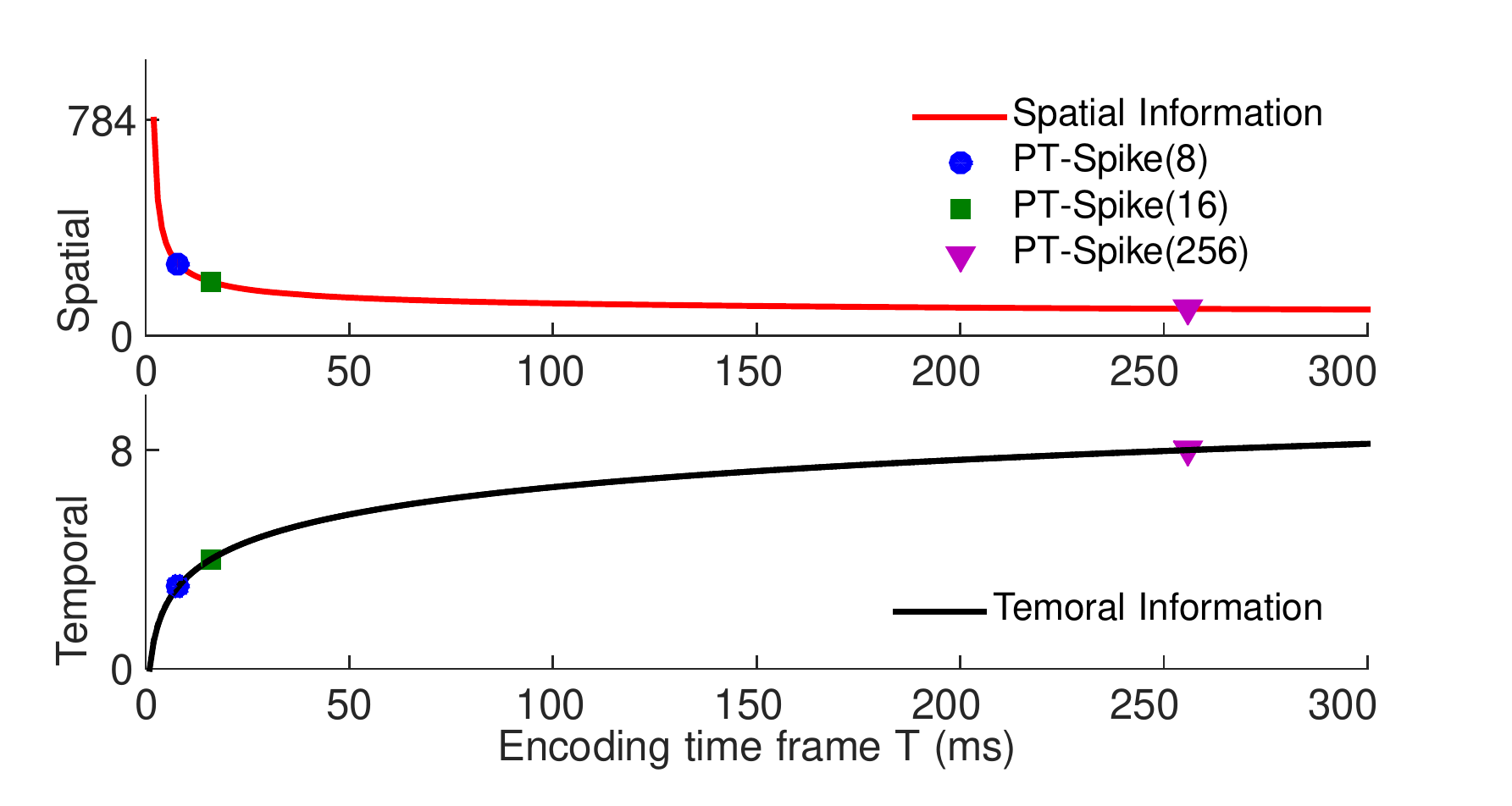}
\par\end{centering}
\vspace{-6pt}
\caption{
\label{st}
Model size reduction through adjustable Temporal Kernel.
}
\end{figure}

Another unique advantage of the proposed ``Precise Temporal Encoding'' is to offer a flexible model size reduction. 
Different from traditional ``one-to-one" mapping, 
various choices of model size reduction can be easily achieved by reconfiguring the size of Temporal Kernel. 
Fig.~\ref{st} illustrates such an interesting concept offered by ``Precise Temporal Encoding''. Increasing the Temporal Kernel size can enrich the temporal information (see encoding time frame from $T=16$ms to $T=256$ms in Fig.~\ref{st}), and hence reduce the needed spatial information or input neurons, e.g. 169 input neurons for ``PT-Spike (16)" v.s. 49 input neurons for ``PT-Spike (256)". The training and inference accuracies will be slightly changed according to the selected Temporal Kernel size (see Section.~\ref{sec:evaluation}). 

\subsection{Synaptic Processing and Linearized Spiking Kernel}
Once the delay for the single spike is determined, as shown in Fig.~\ref{arch}, a spiking kernel ${\mathop{\rm K}\nolimits}$ will be applied to shape the associated spikes for input neurons. The kernel plays an important role in the following synaptic weighting for the output voltage $V_n(t)$, as shown in Eq (~\ref{eq_vv}):
\begin{equation}
\label{eq_vv}
{V_n}(t) = \sum_m^Mw_{mn}\sum_{t_s}^TK(t-t_s)
\end{equation}
where weight $V_n(t)$ represents the voltage of output neuron $n$, $w_{mn}$ denotes the synaptic efficacy between input neuron $X_m$ and output neuron $A_n$. $t_s$ is the decoded spiking delay of $X_m$.
To provide sufficient and accurate temporal information for the classification, the exponential decayed post-synaptic potential in the biological spike response neural model~\cite{gerstner2001framework} can be expressed as:
\begin{equation}
\label{eq_psp}
K_1(t-t_s) = \mu(exp[-(t-{t_s})/\tau_1]-exp[-(t-{t_s})/\tau_2])
\end{equation}
where $\tau$ ($\tau_1$ and $\tau_2$) denotes decay time constant, and $\mu$ is the normalizing constant.
However, such an exponential decaying function requires expensive computation and hardware resource. In ``\textit{PT-Spike}", we employ a more hardware-favorable kernel function ${\mathop{\rm K_2}\nolimits}$--a linear decaying function (see $K_1$ and $K_2$ comparison in Fig.~\ref{arch}), to simplify the costly dual-exponential function $K_1$:
\begin{equation}
\label{eq_k}
K_2(t-t_s) = 1-\tau(t-{t_s})
\end{equation}
As we shall show in Section.~\ref{sec:evaluation}, such a linear approximation cause very marginal classification accuracy degradation. Besides, this linear kernel function will be also applied to detect the input voltage contributions to the output spike in our proposed ``PT-Learning". 

\subsection{Asymmetric Decoding}
In `\textit{PT-Spike}", a novel \underline{A}symmetric decoding scheme, namely ``A-Decoding", is proposed for the classification. As the error signal critical for the proposed supervised temporal learning will be also generated through asymmetric decoding, we will discuss the ``A-Decoding" technique first.

\begin{figure}[b]
\begin{centering}
\includegraphics[width=0.79\columnwidth]{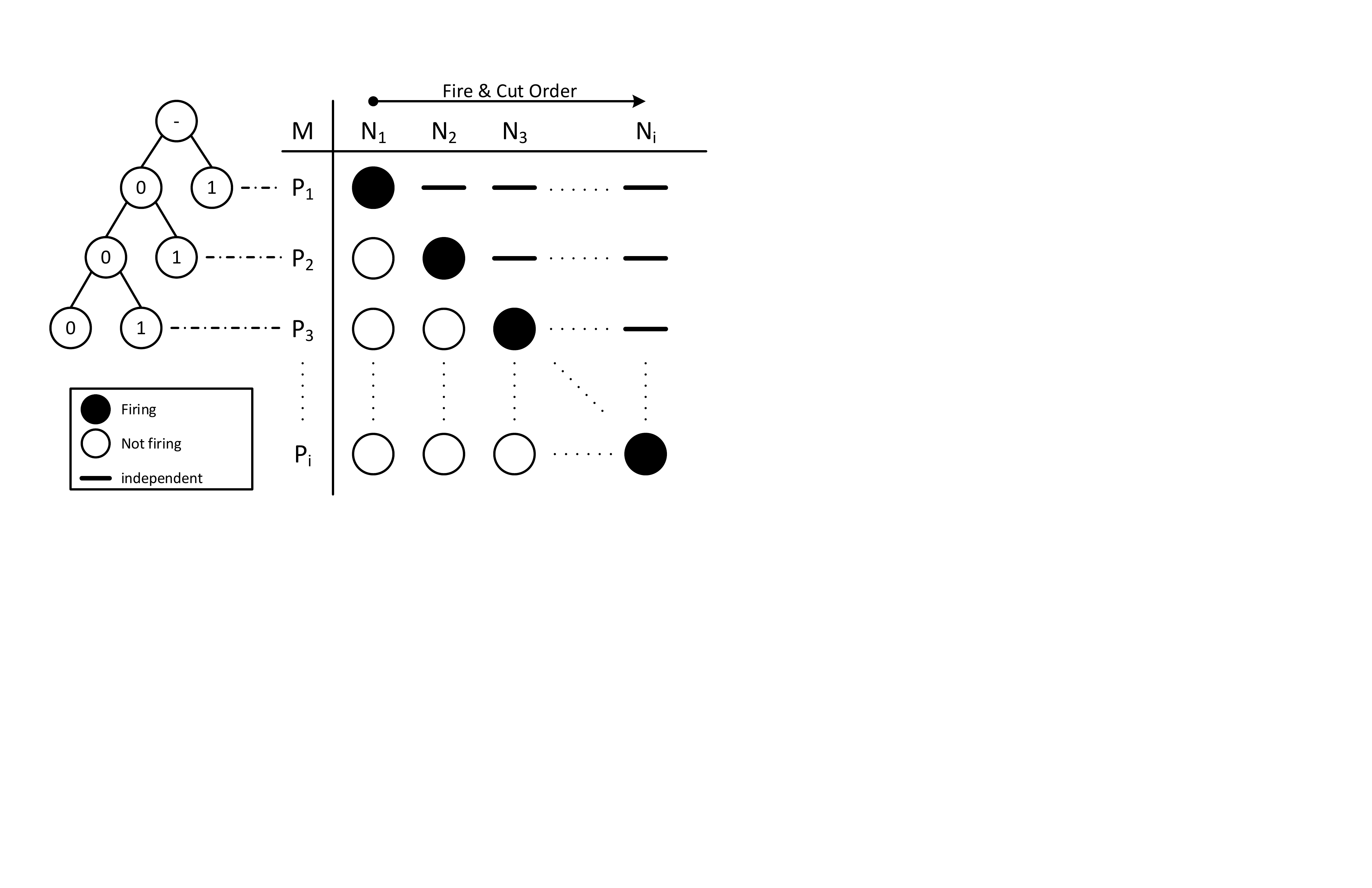}
\par\end{centering}
\caption{
\label{decode}
An overview of proposed ``A-Decoding''.
}
\end{figure}

In rate-based SNN, the target pattern can be determined by the output neuron with highest spiking numbers. 
The costly weight updating will be performed in all synapses at each iteration of learning.
The subsequent neural competition (weight conflict) among different patterns can be rectified by enough information provided by the large number of input spikes. 
Hence a good classification accuracy may be achieved for all different patterns. However, the similar case cannot occur in our proposed ``\textit{PT-Spike}", since its weight updating solely relies on the very limited number of spare spikes (e.g. a single spike) in temporal domain.
In ``\textit{PT-Spike}", we further propose the ``A-Decoding" to alleviate the neural competition for accuracy improvement.

Fig.~\ref{decode} illustrates the key idea of proposed ``A-Decoding", including pattern readout and error detection.
Pattern $\{P_i\}$ can be decoded based on the firing status of output neuron $\{N_i\}$. In our asymmetric decoding, the output neuron can work on three different statuses: ``firing", ``not firing" and ``independent", as shown in Fig.~\ref{decode}. Note ``independent" means that the associated neurons will not participate in the learning process of a certain pattern, and it will only occur in learning mode.

In testing mode, the output neuron will be only in following two status: $\{1-firing/0-not firing\}$.
The target pattern is scanned according to the order of the first firing neuron. Assume a binary code $\tilde{N_1}\tilde{N_2}\tilde{N_3}\cdots\tilde{N_i}$ is generated by output neurons $\{N_i\}$, a Huffman-style decoding procedure can be performed (See Fig.~\ref{decode} left part).
For example, if the first firing neuron is $N_3$, the corresponding code will be $\tilde{0}\tilde{0}\tilde{1}$. Thus, the target pattern is $P_3$. In ``\textit{PT-Spike}", the early detection of testing, namely ``Fire\&Cut'', can be realized based on the temporal ``winner-take-all'' rule: Once the IFC of neuron $N_i$ triggers a spike, all the remained IFCs for other neurons will be shut down by following the ``Fire\&Cut Order'', which may save the additional power consumed by the IFCs.

In learning mode, a desired spike pattern is reversely generated according to the Huffman-style decoding of pattern $\{P_i\}$ (See Fig.~\ref{decode} right part). Once a participated neuron $N_i$ triggers an unexpected firing or a missing firing, an error will be detected and only the synaptic weights of $N_i$ will be modified according to our proposed ``PT-learning''. Note only ``partial'' output neurons (NOT in``independent" status), will be involved during the learning of pattern $\{P_i\}$, namely ``Partial Learning''. Such a mechanism significantly accelerates the learning procedure and saves power consumed by the unnecessary neural processing.
Meanwhile, $\{N_i\}$ is ``asymmetrically'' correlated with $\{P_i\}$ and thus can ease the neural competition. For example, neuron $N_i$ only engages in the synaptic plasticity of pattern $P_i$ and will be ignored during the learning of all other patterns. 
As we shall show later, by taking advantages of ``Fire\&Cut'', ``Partial Learning'' and ``Ease Competition'', our proposed ``A-Decoding'' can significantly enhance the weighting efficiency and learning accuracy.

\newcommand\mycommfont[1]{\footnotesize\textcolor{blue}{#1}}
\SetCommentSty{mycommfont}

\begin{algorithm}[t]
\small
\caption{\label{alg}Post-Synaptic Processing}
\DontPrintSemicolon
\tcp{Pseudocode of Asymmetric Decoding and PT-Learning}

Detecting:\;
\ForEach{output neuron $N_i$ in [$N_1$ .. $N_I$]}{
  \eIf{testing mode}{
    \If{firing}{return $P_i$\tcp{``Fire\&Cut''}
    }
  }{\tcp{learning mode}
    \uIf{$N_i$ is independent to $P_i$}{
      return\tcp{``Partial Learning'' and ``Ease Competition''}
    }
    \ElseIf{actual firing pattern $\neq$ desired pattern}{
      call Learning($V_{max}$, $T_{max}$)\;
    }
  }
}
Learning:\;
\tcp{change synaptic weights of $N_i$}
Err $\leftarrow V_{th} - V_{max}$\;
$T_{fal}$ $\leftarrow T_{max}$\;
\ForEach{input neuron $X_c$ in [$X_1$ .. $X_M$]}{
  \eIf{$K_2(T_{fal}-T_c) \leqslant 0$}{
    continue\tcp{``Partial Updating''}
  }{\tcp{pre-spiking at $T_c$ contributed to post-spiking}
    $\Delta w \leftarrow \lambda Err K_2(T_{fal}-T_c)$\;
    $w_{ci} \leftarrow \Delta w + w_{ci}$\;
  }
}
\end{algorithm}

\subsection{PT-Learning}

Our proposed ``PT-Learning" coordinates with the aforementioned ``A-Decoding" to capture the errors needed for synaptic weights updating.
An error detected by the ``A-Decoding" will be processed by ``PT-Learning'' to generate corresponding weight changes and send back for synapse updating. As shown in Fig.~\ref{arch}, based on the actual and expected spiking pattern, two types of errors may occur in the output neuron: ``false missing" and ``false fire". Here ``false missing" means that the integrated voltage can not reach the threshold in output neuron to trigger the expected output spike, while ``false fire" is defined as an undesired spike firing.

As shown in Algorithm.~\ref{alg}, once an error is detected, the error spiking time ($T_{fal}$) and the cost function ($Err$) will be extracted from $T_{max}$ and $V_{th} - V_{max}$. Here $V_{max}$ and $T_{max}$ are the maximum voltage amplitude and its occurrence time, respectively. A negative (positive) $Err$ means a false- fire (missing). Hence, the gradient of $Err$ with respect to each weight $w_c$ at pre-synaptic spiking time $T_c$ can be calculated as:
\begin{equation}
\label{eq_err}
-\frac{\text{d}Err}{\text{d}w_c}= Err\sum_{T_{c}\leq T_{max}}K_2(T_{max}-T_c)+\frac{\partial V(T_{max})}{\partial T_{max}}\frac{\text{d}T_{max}}{\text{d}w_c}
\end{equation}
Here $K_2$ is the linear decayed spike kernel defined in Eq.(~\ref{eq_k}).

As pre-synaptic spikes are weighted through synaptic efficacy $w_c$ before $T_{max}$, $\frac{\partial V(T_{max})}{\partial T_{max}}$ = 0.
By further considering $Err$ into the change of $w_c$, $\Delta{w_c}$ can be expressed as:
\begin{equation}
\label{eq_dw}
\Delta{w_c} = \lambda Err\sum_{T_{c}\leq T_{fal}}K_2(T_{fal}-T_{c})
\end{equation}
where $\lambda$ denotes the learning rate and spike kernel $K_2$ can be used again to calculate the contributions from the input neuron $X_c$ at time $T_c$.

As discussed in ``A-Decoding", only partial output neurons will be involved during the learning of a certain pattern, meaning that only partial synaptic weights will be updated. The dual-level acceleration, contributed by both ``A-Decoding'' and ``PT-Learning'', can improve the learning efficiency significantly. As we shall show later, the synaptic computation can be reduced more than 200\% when compared with the standard learning approach without accelerations. Moreover, ``PT-Learning" together with ``A-Decoding" can boost the accuracy for realistic recognitions task significantly.


\begin{figure*}[t]
\centering
\begin{subfigure}[t]{0.49\textwidth}
\includegraphics[width=\textwidth]{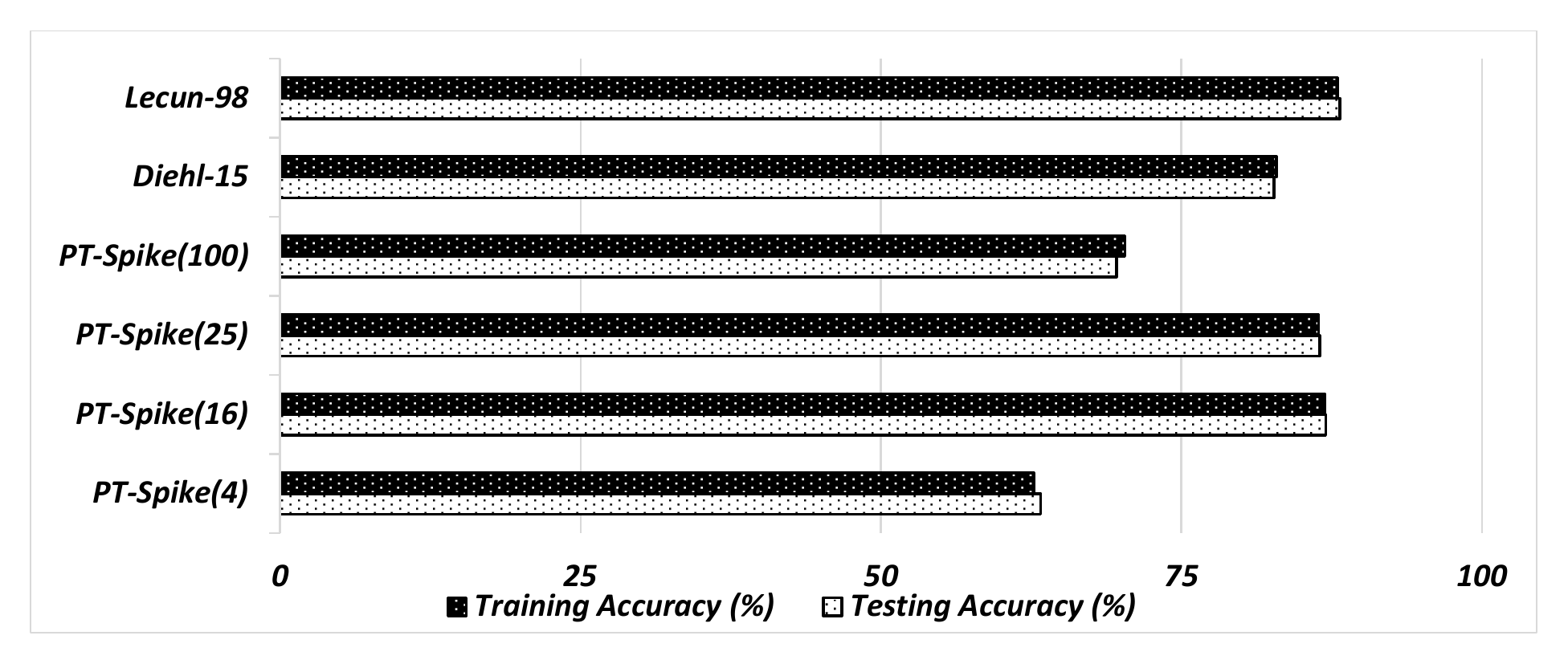}
\caption{\label{e_acc_main}Training and Testing Accuracies of Selected Candidates.}
\end{subfigure}
\begin{subfigure}[t]{0.49\textwidth}
\includegraphics[width=\textwidth]{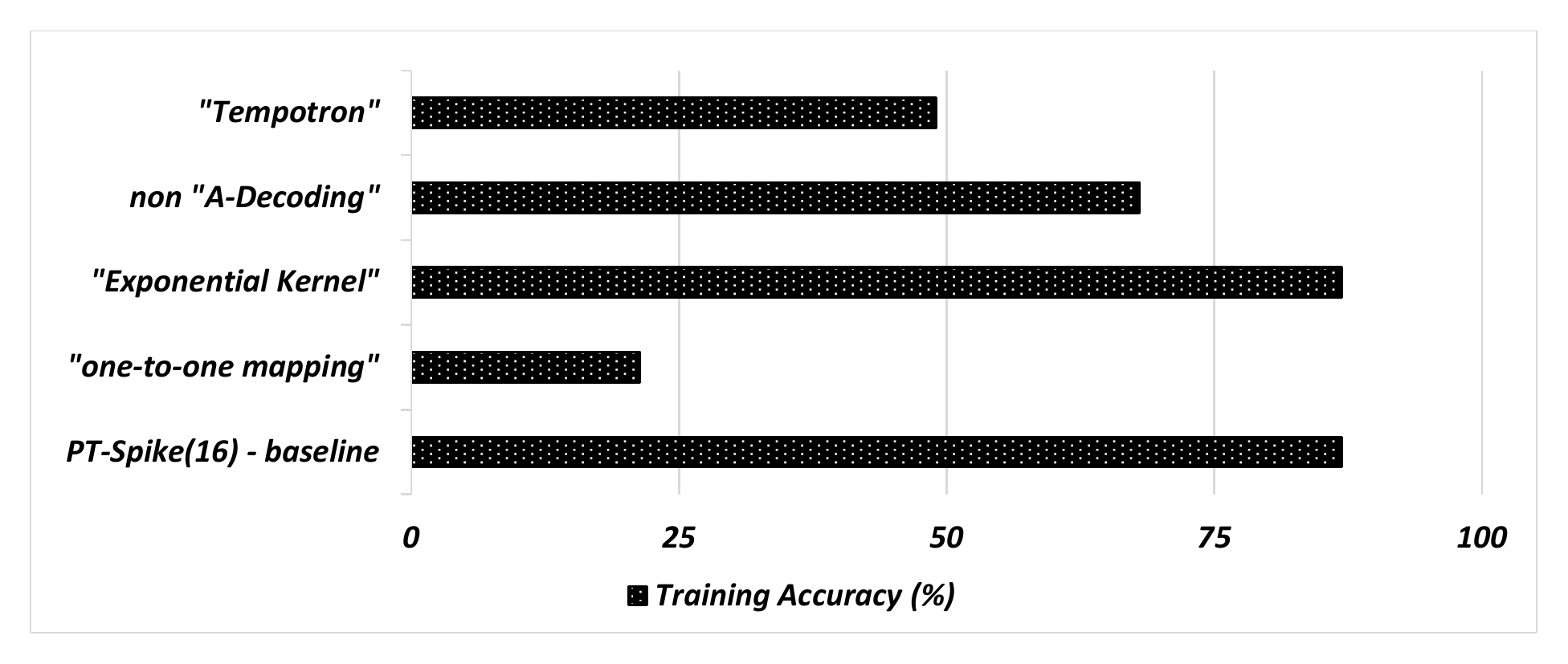}
\caption{\label{e_acc_misc}Training Accuracy with Different Designs.}
\end{subfigure}
\caption{Accuracy Evaluations for Difference Candidates and Design Optimizations.}
\end{figure*}

\begin{figure*}[b]
\centering
\begin{subfigure}{0.3\textwidth}
\includegraphics[width=\textwidth]{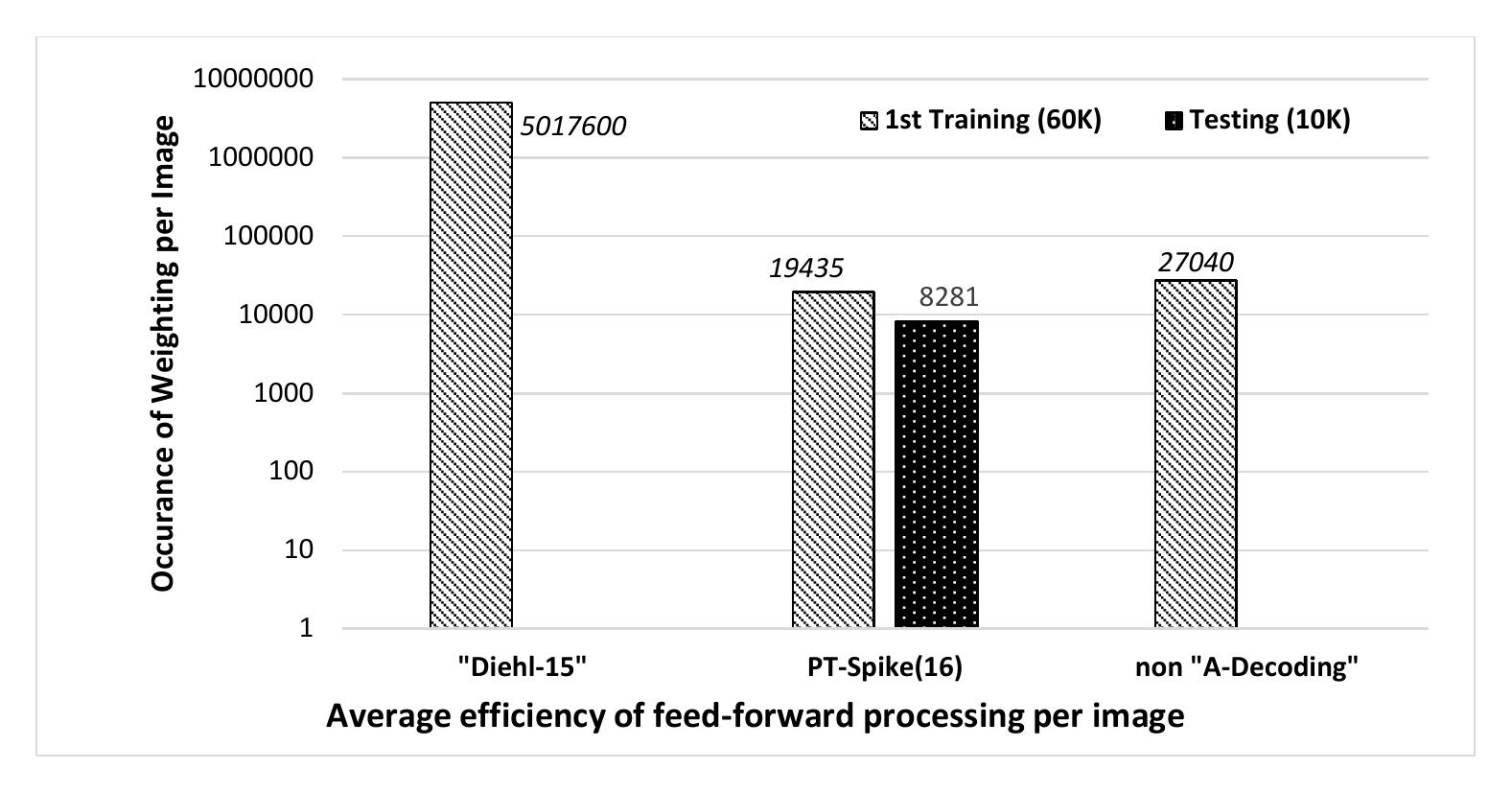}
\caption{\label{e_feedforward_main}Feed-forward Efficiency per Input Image.}
\end{subfigure}
\begin{subfigure}{0.3\textwidth}
\includegraphics[width=\textwidth]{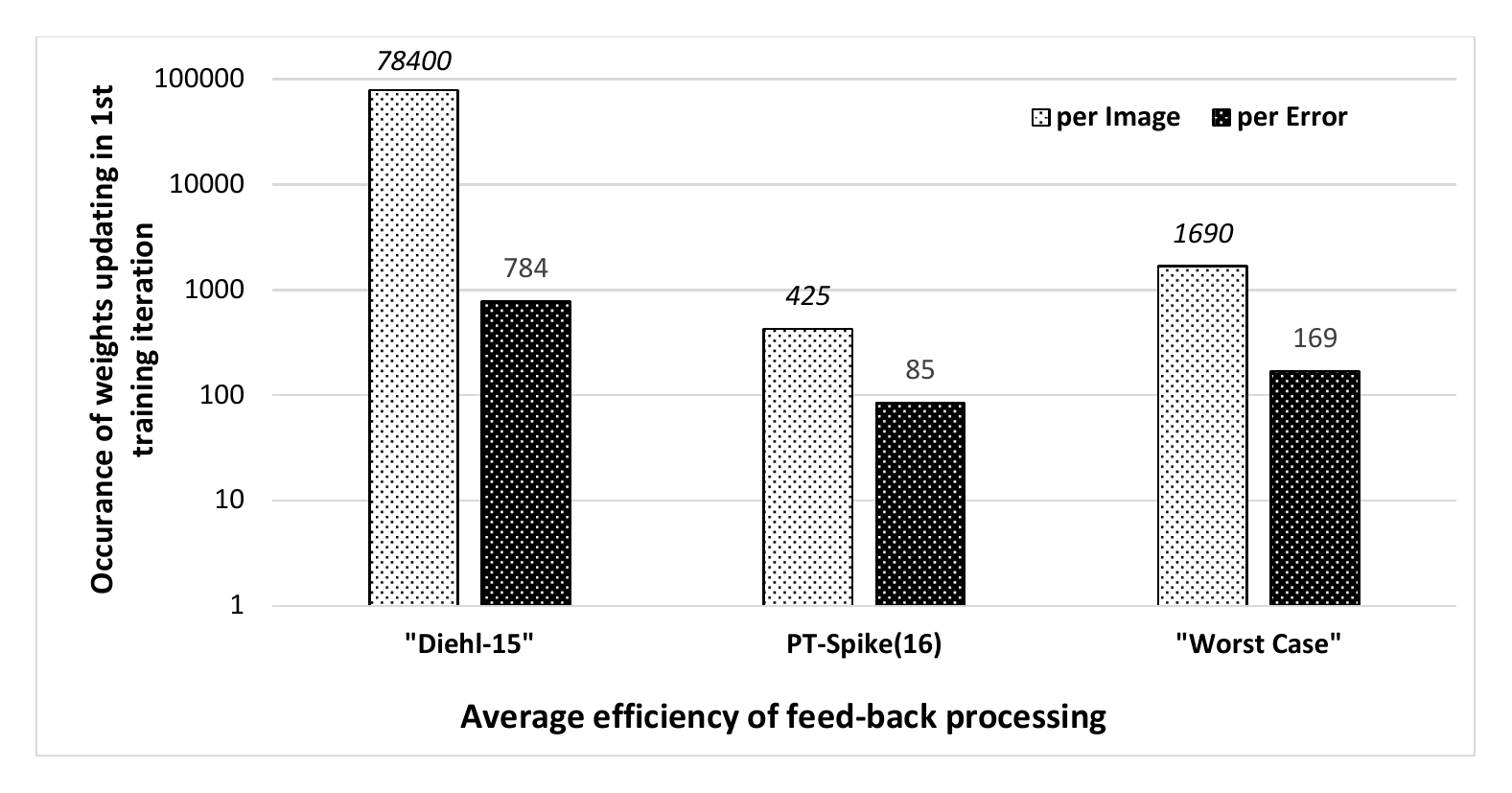}
\caption{\label{e_feedback_main}Feed-back Efficiencies.}
\end{subfigure}
\begin{subfigure}{0.3\textwidth}
\includegraphics[width=\textwidth]{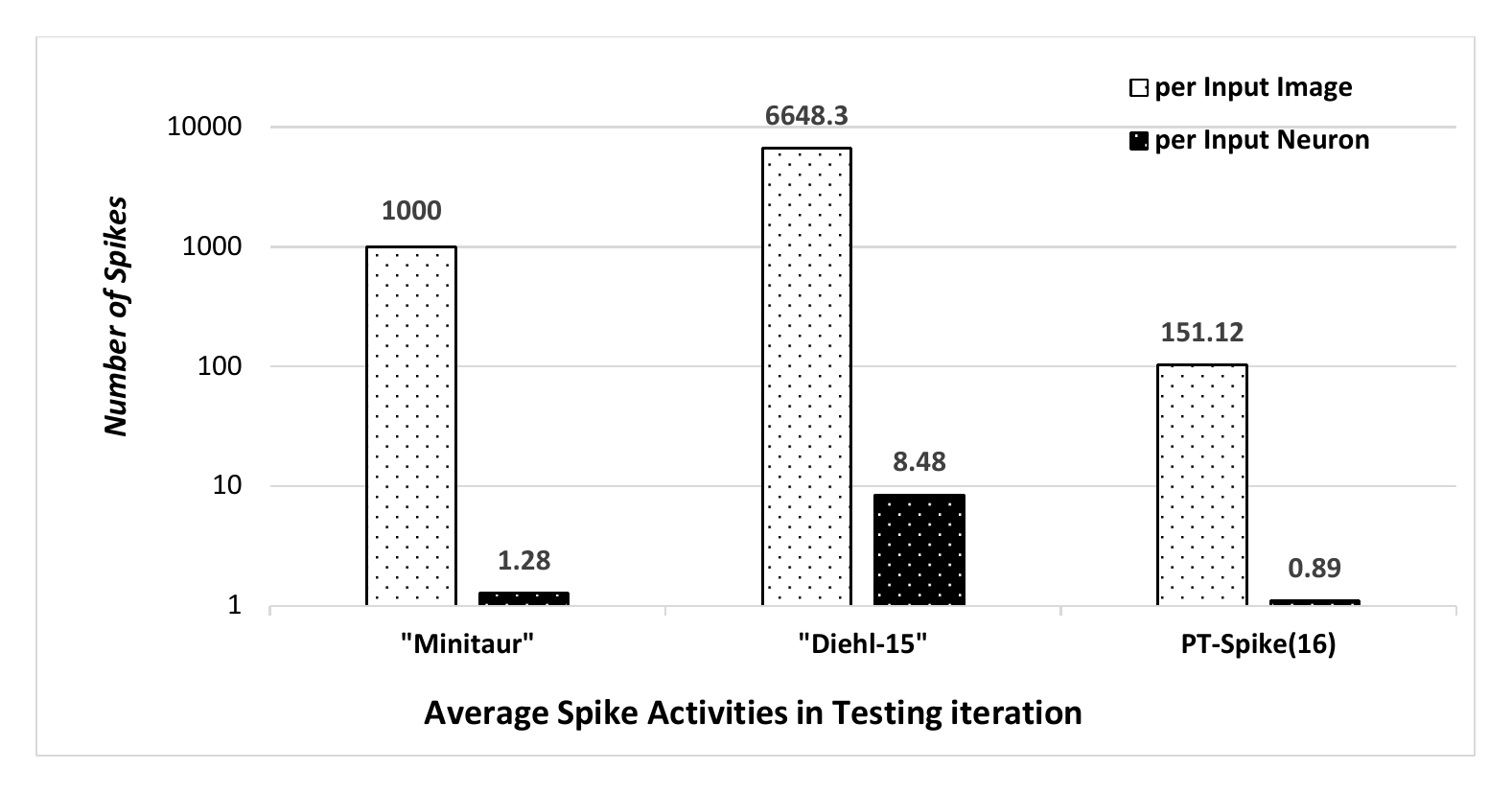}
\caption{\label{e_energy_spikes}Power Consumptions ($\alpha$ Joules $/$spike).}
\end{subfigure}
\caption{Processing Efficiency and Power Consumption}
\end{figure*}

\section{Evaluations}
\label{sec:evaluation}
To evaluate the accuracy, processing efficiency and power consumption of our proposed ``\textit{PT-Spike}" neuromorphic architecture, extensive experiments are conducted in the platforms like MATLAB and heavily modified open-source simulator--Brian~\cite{goodman2009brian}.


\begin{table}[b]
\centering
\caption{Structural Parameters of Selected Candidates.}
\label{tbl_sp}
\resizebox{\columnwidth}{!}{%
\begin{tabular}{|c|c|c|c|c|}
\hline
Candidate & \begin{tabular}[c]{@{}c@{}}Number of \\ input neurons\end{tabular} & \begin{tabular}[c]{@{}c@{}}Number of \\ output neurons\end{tabular} & \begin{tabular}[c]{@{}c@{}}Number of \\ synaptic weights\end{tabular} & \begin{tabular}[c]{@{}c@{}}neural processing \\ time-frame T\end{tabular} \\ \hline\hline
PT-Spike(4) & 196 & 10 & 1960 & 4ms \\ \hline
PT-Spike(16) & 169 & 10 & 1690 & 16ms \\ \hline
PT-Spike(25) & 144 & 10 & 1440 & 25ms \\ \hline
PT-Spike(100) & 100 & 10 & 1000 & 100ms \\ \hline
Diehl-15 & 784 & 100 & 78400 & 500ms \\ \hline
Lecun-98 & 784 & 10 & 7840 & - \\ \hline
\end{tabular}%
}
\end{table}

\subsection{Simulation Setup}
In our evaluation, a full MNIST database is adopted as the benchmark
~\cite{lecun1998mnist}. A set of ``\textit{PT-Spike}" designs--``PT-Spike(R)'' 
are implemented to demonstrate the leveraged temporal encoding where
``R'' denotes the number of interested pixels per input neuron or the size of Temporal Kernel in proposed ``Precise Temporal Encoding''.
We also assume the encoding time frame ($T$) is $T = \tau \times R(ms)$, where $\tau = 1(ms)$ is the fixed minimum time interval to fire the spike. The maximum temporal information $T$ can be adjusted by tuning the parameter $R$. The number of input neurons (spatial domain) can be expressed  as $M = \lceil\frac{P-\sqrt{R}+1}{S}\rceil^{2}$, where $P$ and $S$ represent the width of an input image and the stride with which we slide the Temporal Kernel.
$P=28$ and $S=2$ are selected in our evaluations of MNIST dataset. 
Two representative baselines under similar network configurations, including the rate-coded SNN--``Diehl-15''~\cite{diehl2015unsupervised} and the ANN--``Lecun-98''~\cite{lecun1998gradient}, are also implemented for the energy and performance comparisons with proposed ``\textit{PT-Spike}". 

Table.~\ref{tbl_sp} presents the detailed structural parameters of selected candidates. Compared with the ``Diehl-15'' and ``Lecun-98'', our proposed temporal encoding achieves significant model size reduction for all ``PT-Spike" designs, i.e. $\sim40\times$ (``PT-Spike(4)" v.s. ``Diehl-15'') and  $\sim4\times$ (``PT-Spike(4)" v.s. ``Lecun-98'').



\subsection{Accuracy}

Fig.~\ref{e_acc_main} shows the accuracy comparison among different ``PT-Spike (R)'', ``Lecun-98'' and ``Diehl-15''. ``PT-Spike(25)'' can achieve very comparable accuracy at much lower cost ($\sim86$\%, 1440 synaptic weights) when compared with ``Diehl-15'' ($\sim83$\%, 78400 synaptic weights) and ``Lecun-98'' ($\sim88$\%, 7840 synaptic weights). Meanwhile, ``PT-Spike(16)'' and `PT-Spike(25)'' also show a very close accuracy ($\sim87$\% and $\sim86$\%), which is much better than ``PT-Spike(4)'' and ``PT-Spike(100)'' ($\sim63$\% and $\sim70$\%). 



We also evaluated the individual training accuracy improvement contributed by various proposed techniques, such as ``linearized spiking kernel'', ``Precise Temporal Encoding'', ``A-Decoding" and ``PT-Learning", receptively. Here, we choose the ``PT-Spike(16)" as the baseline design that employs all aforementioned techniques.  ``Exponential Kernel'', ``one-to-one mapping'', ``non A-Decoding'' and ``Tempotron" denote the designs that substitute only one out of the four techniques.
As shown in Fig.~\ref{e_acc_misc}, ``PT-Spike(16)" shows a very marginal accuracy degradation ($0.2$\%) because of the ``linearized spiking kernel'' ($K_2$ in Eq.(~\ref{eq_k})) when compared with the original costly ``Exponential Kernel'' design ($86.9$\%, $K_1$ in Eq.(~\ref{eq_psp})). Furthermore, ``PT-Spike(16)'' boosts the accuracy by $\sim400$\%, $\sim19$\% and $\sim38$\% when compared with the designs of ``one-to-one mapping'' ($\sim21$\%), ``non A-Decoding'' ($\sim68$\%), and the theoretical ``Tempotron'' learning rule ($\sim49$\%), respectively, which clearly demonstrates the effectiveness of the proposed ``Precise Temporal Encoding'', ``A-Decoding'' and ``PT-Learning''.

\subsection{Processing Efficiency}

The occurrence frequency of synaptic events is calculated to evaluate the system processing efficiency, including both weighting and weights updating. 
Fig.~\ref{e_feedforward_main} compares the number of weighting operations among three designs in the feed-forward pass. 
Unlike the other candidates, the amount of weight operations of ``PT-Spike(16)" is different between training and testing due to the ``Fire\&Cut" mechanism in``A-Decoding". Hence, the weighting of the first testing iteration is also included in ``PT-Spike(16)". Even the ``non A-Decoding'', i.e. ``PT-Spike(16)" without the ``A-Decoding'' technique, gains $\sim 185\times$ weighting operation reduction as compared with ``Diehl-15'' since rate-coded SNN requires a long time window to process the spikes with enlarged neuron model size,
causing tremendous weighting processes on each time slot. Compared with ``non A-Decoding'', weighting operations of ``PT-Spike(16)'' can be further reduced by $\sim28$\% and $\sim69$\% in first training iteration and testing iteration, respectively. As expected, the ``early-detection'' working mechanism in ``A-Decoding" removes many unnecessary weighting operations on both ``initialized'' weights and ``well-trained'' weights. 

We also characterize the occurrence frequency of weights updating during the first training iteration to evaluate the processing efficiency in the feed-back pass. 
As Fig.~\ref{e_feedback_main} shows, even ``Worst Case'' (i.e. ``PT-Spike(16)'' without employing ``A-Decoding'' and ``PT-Learning'') achieves $\sim4.6\times$ and $40\times$ reductions on weights updating per image and per error, respectively, when compared  with ``Diehl-15''. Such impressive improvement is introduced by the significant compressed model size. Moreover, compared with the ``worst case'', ``PT-Learning'' and ``A-Decoding" contribute $\sim2\times$ and $\sim4\times$ weights updating reduction per error and per image for ``PT-Spike(16)'', respectively, demonstrating the effectiveness of ``dual-level acceleration'' from decoding and learning. 

\subsection{Power Consumption}
To roughly evaluate the power efficiency contributed by the proposed architecture, we adopted a similar methodology used in ~\cite{akopyan2015truenorth,cao2015spiking}. 
A new candidate ``Minitaur"~\cite{neil2014minitaur} is introduced for a fair comparison since it is a more hardware-oriented rate-coded SNN. As Fig.~\ref{e_energy_spikes} shows, ``PT-Spike(16)" saves $\sim8\times$ and $\sim64\times$ power for each input neuron and each input image over ``Diehl-15", respectively, indicating the efficiency of our proposed single-spike coding technique. Compared with the hardware-oriented rate-coded SNN design ``Minitaur", ``PT-Spike(16)" can still achieve $\sim1.4\times$ ($\sim6.6\times$) power reduction on each input neuron (input image). 



\subsection{Discussions}
The research of the time-based SNN represented by extreme sparse spikes, i.e. \textbf{single spike design}, is still in its infancy, and to our best knowledge, we have not seen any exemplar large networks successfully demonstrated for performing the realistic cognitive tasks. Due to the 
unique time-based learning and information representation, the research in this area is quite challenge and unique. 
In this work, we adopt a proof-of-concept simple design, i.e. Single-Layer Perceptron to illustrate the design optimizations of the time-based SNN, and demonstrate its potentials for realistic applications,
though the classification accuracy is still lower than that of state-of-the art DNNs and CNNs.

Extending our design to multi-layered network will enhance its capability to handle more complicated cognitive tasks, however, is non-trivial, as a multi-layer learning rule needs to be developed to facilitate the spatial information transfer among different layers. While our proposed approach cannot be directly applied for the multi-layered network in its current form, the novel techniques proposed in this paper, i.e. ``Temporal Kernel Coding", ``PT-Learning" and ``A-Decoding" form the basis for the time-based multi-layer network.
We believe the initial architecture developed in this paper will serve as a basic framework to the multi-layer network design, and may encourage more interesting researches in this domain.


\section{Conclusion}
\label{sec:conclusion}
As the rate-based spiking neural network (SNN) is subject to power and speed challenges due to processing large number of spikes, in this work, 
we systematically studied the possibility of utilizing the more power-efficient time-based SNN in real-world cognitive tasks. 
Three integrated techniques--precise temporal encoding, efficient supervised temporal learning and fast asymmetric decoding, were proposed to construct the Precise-Time-Dependent Single Spike Neuromorphic Architecture, namely, \textit{``PT-Spike"}. The single-spike temporal encoding offers an energy-efficient information 
representation solution with the potentials of 
model size reduction.
The supervised learning and asymmetric decoding can work cooperatively to deliver a more effective and efficient synaptic weight updating and classification. Our evaluations on the MNIST database well demonstrate the advantages of \textit{``PT-Spike"} over the rate-based SNN in terms of network size, speed and power, with a comparable accuracy. 

\footnotesize
\bibliographystyle{IEEEtran}
\bibliography{cites,Reference,NN,spiking}

\end{document}